\documentclass[10pt,twocolumn,letterpaper]{article}

\usepackage{iccv}
\usepackage[accsupp]{axessibility}  
\usepackage{times}
\usepackage{epsfig}
\usepackage{graphicx}
\usepackage{amsmath}
\usepackage{amssymb}
\usepackage[format=plain,labelformat=simple,labelsep=period,font=small,compatibility=false]{caption}
\usepackage[font=footnotesize,skip=3pt,subrefformat=parens]{subcaption}
\usepackage{float}

\usepackage{booktabs}
\usepackage[table]{xcolor}
\usepackage{bm}
\usepackage{nicefrac}
\usepackage{braket}
\newlength\savewidth\newcommand\shline{\noalign{\global\savewidth\arrayrulewidth
  \global\arrayrulewidth 1pt}\hline\noalign{\global\arrayrulewidth\savewidth}}
\newcommand{\tablestyle}[2]{\setlength{\tabcolsep}{#1}\renewcommand{\arraystretch}{#2}\centering\footnotesize}

\newcommand{\rot}[1]{\rotatebox[origin=lb]{90}{\smash{#1}}}

\newcolumntype{x}[1]{>{\centering\arraybackslash}p{#1pt}}
\newcolumntype{y}[1]{>{\raggedright\arraybackslash}p{#1pt}}
\newcolumntype{z}[1]{>{\raggedleft\arraybackslash}p{#1pt}}

\definecolor{baselinecolor}{gray}{.9}
\newcommand{\baseline}[1]{\cellcolor{baselinecolor}{#1}}
\definecolor{tsneRed}{rgb}{0.85, 0.13, 0.16}
\newcommand{\red}[1]{\textcolor{tsneRed}{#1}}
\definecolor{tsneGreen}{rgb}{0.00, 0.59, 0.20}
\newcommand{\gre}[1]{\textcolor{tsneGreen}{#1}}
\definecolor{before}{RGB}{191,191,225}
\definecolor{avg}{RGB}{225, 191, 192}
\definecolor{last}{RGB}{240,204,126}
\newcommand{\Gb}{\cellcolor{before}}
\newcommand{\Ga}{\cellcolor{avg}}
\newcommand{\Gl}{\cellcolor{last}}

\usepackage[pagebackref=true,breaklinks=true,letterpaper=true,colorlinks,bookmarks=false]{hyperref}

\usepackage[capitalize]{cleveref}
\crefname{section}{Sec.}{Secs.}
\Crefname{section}{Section}{Sections}
\Crefname{table}{Table}{Tables}
\crefname{table}{Tab.}{Tabs.}

\iccvfinalcopy 

\def\iccvPaperID{6658} 

\ificcvfinal\fi

\begin{document}

\title{Preventing Zero-Shot Transfer Degradation in Continual Learning of Vision-Language Models}

\author{
Zangwei Zheng$^{1}$\quad Mingyuan Ma$^{2}$\quad Kai Wang$^{1}$\quad Ziheng Qin$^{1}$\quad Xiangyu Yue$^{3}$\quad Yang You$^{1}$
\vspace{1mm}
\\
$^1$National University of Singapore\quad $^2$UC Berkeley\quad $^3$The Chinese University of Hong Kong
\vspace{2mm}
\\ 
{\normalsize \textsuperscript{\rm 1}\{zangwei, kai.wang, zihengq, youy\}@comp.nus.edu.sg}
{\normalsize \textsuperscript{\rm 2}mamingyuan2001@berkeley.edu}
{\normalsize \textsuperscript{\rm 3}xyyue@ie.cuhk.edu.hk}
}

\maketitle
\ificcvfinal\thispagestyle{empty}\fi

\begin{abstract}
Continual learning (CL) can help pre-trained vision-language models efficiently adapt to new or under-trained data distributions without re-training. Nevertheless, during the continual training of the Contrastive Language-Image Pre-training (CLIP) model, we observe that the model's zero-shot transfer ability significantly degrades due to catastrophic forgetting.
Existing CL methods can mitigate forgetting by replaying previous data. However, since the CLIP dataset is private, replay methods cannot access the pre-training dataset. In addition, replaying data of previously learned downstream tasks can enhance their performance but comes at the cost of sacrificing zero-shot performance.
To address this challenge, we propose a novel method ZSCL to prevent zero-shot transfer degradation in the continual learning of vision-language models in both feature and parameter space. 
In the feature space, a reference dataset is introduced for distillation between the current and initial models. The reference dataset should have semantic diversity but no need to be labeled, seen in pre-training, or matched image-text pairs. 
In parameter space, we prevent a large parameter shift by averaging weights during the training.
We propose a more challenging Multi-domain Task Incremental Learning (MTIL) benchmark to evaluate different methods, where tasks are from various domains instead of class-separated in a single dataset. Our method outperforms other methods in the traditional class-incremental learning setting and the MTIL by 9.7\% average score. Our code locates at \url{https://github.com/Thunderbeee/ZSCL}.
\end{abstract}

\section{Introduction}
\label{sec:intro}

\begin{figure}[t]
    \centering
    \includegraphics[width=\columnwidth]{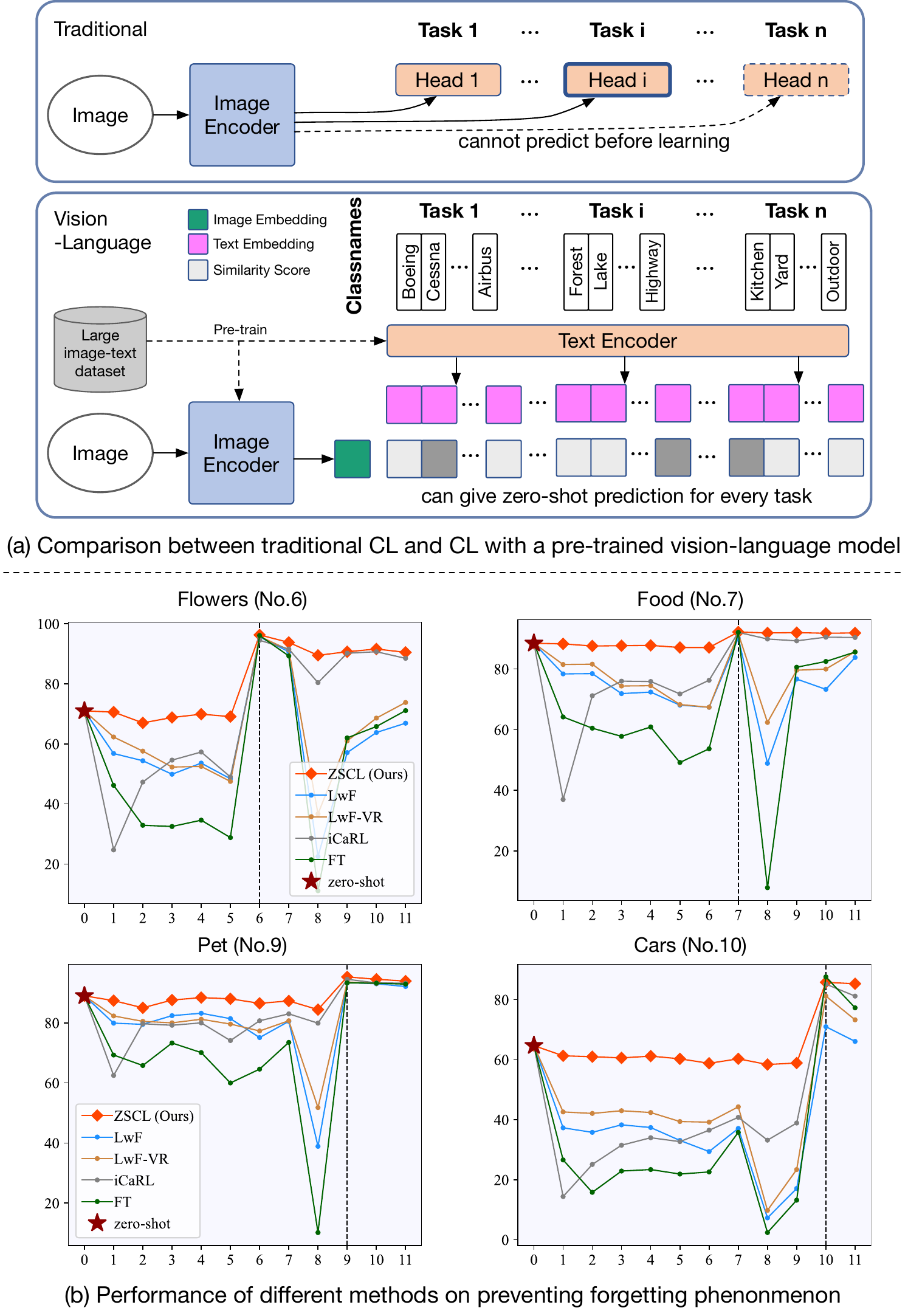}
    \caption{a) Conventional CL learns distinct task-specific heads, while CL with vision-language models can predict both learned tasks and out-of-distribution tasks. b) Accuracy (\%) changes during CL of four datasets on 11 datasets. Our method is superior to others in preventing the forgetting of both zero-shot transfer ability and new knowledge.}
    \label{fig:head}
\end{figure}

Most deep learning models can access all the data during training~\cite{krizhevsky2017imagenet,he2016deep,dosovitskiy2020image}.
If we want to expand a model's knowledge, such as learning a newly found animal species~\cite{Parkhi2012CatsAD}, we can re-train the model by adding new classes to the training dataset. However, re-training a large model is costly.
In contrast, continual learning (CL)~\cite{li2017learning,rebuffi2017icarl} incrementally learns task one after another.
It can reduce this cost by only learning the new data, thereby presenting itself as an efficient alternative to conventional learning methods~\cite{qin2023infobatch,liu2023dream,zhou2023dataset}.
Nonetheless, a model tends to forget previous information catastrophically when learning new tasks~\cite{rebuffi2017icarl,li2017learning,wu2019large}. 
The ``catastrophic forgetting'' phenomenon is a great challenge for CL.

Recently, vision-language models have shown powerful zero-shot transfer ability~\cite{radford2021learning, jia2021scaling, Li2022BLIPBL}. They can give zero-shot predictions without any training examples of a task.
However, the performance on some tasks is poor due to insufficient relevant image-text pairs in the pre-training datasets. For example, it is difficult for CLIP~\cite{radford2021learning} to distinguish among digital numbers, with an accuracy on MNIST~\cite{deng2012mnist} below 60\%
much lower than a naively trained CNN~\cite{lecun1998gradient}. If we want to widen the knowledge in the vision-language model by re-training, the computational cost is too large (e.g., CLIP is pretrained on 400 million image-text pairs). Fine-tuning downstream tasks achieves high performance, but one model for a task takes much memory, and the model is not reusable.
Prompt learning~\cite{zhou2022learning,zhou2022conditional} keeps the backbone parameters unchanged. However, it is only effective with limited training data due to a limited prompt length~\cite{zhou2022learning,jin2021good}.
In contrast, continual learning makes learning new knowledge a lifelong process for the vision-language model.
The continually learned model can handle any image-text input and can be further used for downstream tasks~\cite{ding2022don,thengane2022clip}. 

We find that existing CL methods hardly prevent the forgetting phenomenon for zero-shot transfer ability in continual learning of a pre-trained vision-language model. As shown in \cref{fig:head} (a), the CL with a pre-trained vision-language model differs from the traditional one. Besides forgetting previously learned task knowledge, the CLIP-based CL suffers from forgetting pre-training knowledge, namely a degradation of zero-shot transfer ability. For the replay-based CL methods~\cite{rebuffi2017icarl,shin2017continual,lopez2017gradient,isele2018selective,lavda2018continual,prabhu2020gdumb}, the dataset during pre-training may be private and inaccessible during fine-tuning. For distillation-based CL methods~\cite{li2017learning,dhar2019learning,douillard2020podnet,ding2022don}, they do not lay enough emphasis on the pre-trained model. On the one hand, a large model state change hinders tasks thereafter from using high-quality feature representations. On the other hand, it significantly degrades zero-shot performance on unseen datasets.

Our method ZSCL protects the Zero-Shot transfer ability during Continual Learning. We view the knowledge stored in the pre-training model from two perspectives: a well-learned feature space and a good value in the parameter space. In feature space, we re-design previous distillation loss~\cite{hinton2015distilling, rebuffi2017icarl} with different loss styles, teacher models, and data sources. We find the original CLIP model, as opposed to the newly acquired model, is the best option for the teacher model. Instead of using data collected from previous tasks~\cite{rebuffi2017icarl} or current task~\cite{hinton2015distilling}, we find a reference dataset with diverse semantics (e.g., images sampled from ImageNet) is a good option for distillation loss. The reference images need not be labeled or matched with the text. Preserving the relative similarity between reference images and texts makes the feature space deviate little from the original. In the parameter space, WiSE-FT~\cite{wortsman2022robust} proposes interpolating the initial and fine-tuned model for better performance. Inspired by this, we ensemble the weights throughout continuous training to prevent a significant shift from the initial CLIP, which can be seen as interpolating models of different zero-shot transfer and downstream task performance tradeoffs. The weight ensemble method is more stable and not sensitive to hyper-parameters.

To better evaluate our method, we propose a new benchmark Multi-domain Task Incremental Learning (MTIL). Previous CL tasks are crafted by separating classes in one dataset~\cite{douillard2022dytox,yan2021dynamically,zhu2021prototype}, where the images and classes are within a single domain. In contrast, MTIL consists of data from different sources requiring different expert knowledge. It comprises 11 tasks ranging from animal species to aircraft series recognition. As displayed in \cref{fig:head} (b), when sequentially training CLIP on 11 datasets, the drop in the performance of task $i$ after training task $i$ is the traditional forgetting phenomenon. The degradation in the accuracy compared to the original zero-shot one before training task $i$ represents the forgetting in zero-shot transfer ability. Our method better protects the zero-shot transfer ability and preserves the learned knowledge. We outperform previous methods in both conventional class-incremental learning and MTIL settings.
In \cref{fig:head} (b), 

To summarize, our contributions are as follows:
\begin{itemize}
    \item We investigate continual training with the vision-language model and demonstrate the importance of preserving zero-shot transfer ability. A more challenging benchmark MTIL is proposed to evaluate CL methods where the tasks come from distinct domains.
    \item We propose a novel method ZSCL to mitigate the catastrophic forgetting problem in continual learning of the vision-language model by distillation in the feature space and weight ensemble in the parameter space. 
    \item The proposed ZSCL outperforms all state-of-the-art methods across multiple benchmark datasets. On 10 steps CL of CIFAR100 and TinyImageNet, our method outperforms the best of previous ones by 7.7\% and 6.0\% for the Last accuracy. On MTCL, ZSCL outperforms others by 10.9\% on Transfer and 9.7\% on Avg. scores.
\end{itemize}
\section{Related Work}
\label{sec:related_work}

\paragraph{Vision-Language Models.}
Inspired by the success of language foundation models such as GPT-3~\cite{brown2020language} and T5~\cite{raffel2020exploring}, a series of work pre-train vision-language models on large-scale image-text datasets~\cite{lei2021less,radford2021learning,jia2021scaling}. Among them, Contrastive Language-Vision Pre-training~\cite{radford2021learning} achieves remarkable performance on various downstream tasks. It concentrates on aligning images and texts to acquire a joint embedding space.
The CLIP model contains an image encoder~\cite{he2016deep,dosovitskiy2020image} and a text encoder~\cite{devlin2018bert}. During pre-training, contrastive learning is performed in which a paired image-text is a positive pair while image and text from different image-text pairs form a negative pair. 
For inference, the closest text embedding for the image is chosen as the prediction. Vision-language models can give zero-shot predictions on unseen tasks with a robust zero-shot transfer ability on various downstream tasks.

\paragraph{Continual Learning Methods.} Most existing continual learning methods can be categorized into four groups: parameter expansion, memory replay, distillation loss, and parameter regularization. Parameter expansion methods such as DyTox~\cite{douillard2022dytox} and DEN~\cite{yoon2017lifelong} 
introduce new parameters for new tasks. As we want to achieve a more powerful CLIP model at the end of CL, we do not change the architecture of the CLIP model. Memory replay methods~\cite{shin2017continual,lopez2017gradient,prabhu2020gdumb,lavda2018continual,ostapenko2022continual} including iCaRL~\cite{rebuffi2017icarl} and SER~\cite{isele2018selective} keep a memory for exemplars from previously learned tasks. However, pre-training datasets are too large for choosing exemplars or may not be available at downstream training, and downstream data are not good exemplars for preserving the pre-training knowledge. Distillation loss such as LwF~\cite{li2017learning}, LwM~\cite{dhar2019learning}, LwF-VR~\cite{ding2022don}, and PodNet~\cite{douillard2020podnet} aligns current output space with previous ones, whereas distillation based on current tasks are not strong enough to maintain foundational knowledge. For the CLIP model, we find that general images, even if never seen by the model, plus sentences with enough semantics, can be a good ``replay'' for the distillation loss. The parameter regularization loss restricts the flexibility of model parameters by training loss~\cite{kirkpatrick2017overcoming,zenke2017continual,aljundi2018memory} or weight averaging~\cite{lee2017overcoming,wortsman2022robust}. Although this type of strategy performs badly compared to other ways in previous research~\cite{van2019three,belouadah2021comprehensive}, we found it helpful for CLIP continual learning. The limited parameter space prevents the model from diverging significantly from its original state.

\paragraph{Vision-Language Models for Downstream Tasks.}
Many works propose different training strategies of vision-language models for better performance on downstream tasks, such as CoOp~\cite{zhou2022learning}, CLIP-Adapter~\cite{gao2021clip} and WiSE-FT~\cite{wortsman2022robust}. However, very few attempts at continual learning exist. While \cite{srinivasan2022climb,wang2022continual} focus on CL in the pretrain of VL and \cite{ni2023continual} trains VL from scratch using a small dataset, our problem setting differs as we address CL in downstream tasks with a pretrained VL. Currently, most VL models are trained directly on an accessible large dataset, yet integrating knowledge continuously into a pretrained VL is a practical necessity. None of these studies address the zero-shot transfer degradation phenomenon, instead focusing on the traditional forgetting of learned knowledge. Recently, Thengane~\etal~\cite{thengane2022clip} shows CLIP zero-shot prediction achieves state-of-the-art performance in CL settings even without any training. LwF-VR~\cite{ding2022don} is a modified LwF method for the CLIP model where random-generated sentences are used for distillation loss. However, it only considers the feature space, and the distillation with random sentences cannot protect the vision backbone. Differently, we re-examine what should be used for distillation in the feature space and combine the parameter space weight ensemble to provide better performance for the vision-language model continual learning.
\section{Approach}
\label{sec:approach}

\begin{figure*}[t]
    \centering
    \includegraphics[width=\textwidth]{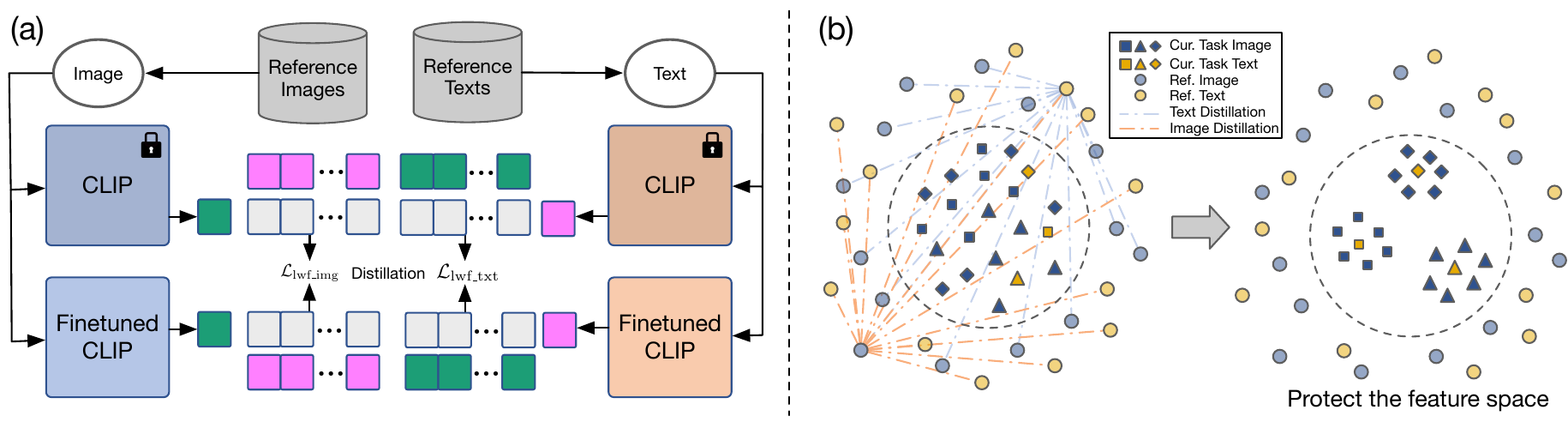}
    \caption{Illustration of ZSCL in feature space. Fig. 3(a) shows the pipeline of distillation. The original and the current model encode the reference images and texts, respectively. The probability distribution of images and texts with respect to each other is distilled. Fig. 3(b) displays how distillation loss preserves the feature space. Compared with the reference dataset, the features of fine-tuning tasks lie in a small subspace. The distillation loss preserves the structure of the feature space by maintaining relative distances.}
    \label{fig:framework}
\end{figure*}

\subsection{Preliminaries}

\paragraph{Continual Learning.} Given $n$ tasks $[\mathcal{T}^1, \mathcal{T}^2, \cdots, \mathcal{T}^n]$, continual training is conducted in sequence on each task $\mathcal{T}^i=(\mathcal{D}^i, C^i), i=1,\dots,n$. Here, $\mathcal{D}^i$ represents the task dataset $\{(\bm{x}^i_j, \bm{y}^i_j\}_{j=1}^{N_i}$, where $\bm{x}^i_j$ is an image, $\bm{y}^i_j$ is a one-hot vector indicating the ground truth, and $N_i$ is the number of images in the dataset. Class names $C^i=\{c^i_j\}_{j=1}^{m_i}$ maps the label of an image to an object name, where $m_i$ is the number of classes for task $\mathcal{T}^i$. The objective of continual learning is to achieve good performance on all tasks. 

Two continual learning settings are focused on in this study~\cite{van2019three}. In task-incremental learning, at inference, the image $\bm{x}$ to be predicted is given with its task identity $t$, so the model only needs to distinguish between different classes in $C^t$. In class-incremental learning, the task identity $t$ is not given. Thus we need to predict with the combined class set $C=\bigcup_{i=1}^nC^i$.

\paragraph{CLIP model.} The CLIP model contains an image encoder $f_i$ and a text encoder $f_t$. The inference process of the CLIP model for image classification is as follows. First, for task $\mathcal{T}^i$, each class $c$ is transformed into a sentence by a template like ``a photo of \{$c$\}''. Then $f_t$ encodes the classes into text embeddings $\{\bm{t}^i_j\}_{j=1}^{m_i}$. An image encoder encodes input image $\bm{x}_k$. The similarity score between the image embedding and text embeddings are calculated as $\bm{s}^i_{k,j}=\Braket{f_i(\bm{x}_k),\bm{t}^i_j}$, where $\Braket{\cdot,\cdot}$ denotes the cosine similarity. The class with the largest similarity score is the prediction for the image. 

To fine-tune the CLIP model for downstream tasks, cross-entropy loss $\text{CE}$ is applied to the similarity score with a temperature scaling:
\begin{equation}
    \mathcal{L}_{\text{CE}} = \frac{1}{N}\sum_{i=1}^N\text{CE}(\tau\cdot \bm{s}_i, \bm{y}_i),
\end{equation}
where $\tau$ is a parameter learned during the pre-training.

\subsection{Distillation in Feature Space}

Well-learned feature space for aligned images and texts enables vision-language models' strong zero-shot transfer ability. It also facilitates the learning of downstream tasks. Compared with the pre-training dataset, downstream datasets lie in a small scope in the feature space (shown in \cref{fig:framework}(b)). Direct fine-tuning of downstream tasks greatly distorts the feature distribution of out-of-distribution data, which leads to a significant drop in zero-shot prediction performance.

While the cross-entropy improves the performance by altering fine-tuned feature subspace, we need a new regularization to preserve the potential out-of-distribution feature space. The relative similarity between one image and different texts is:
\begin{equation}
    \bm{p}=\text{Softmax}(\bm{s}_1, \cdots, \bm{s}_{m}).
\end{equation}
We hope the above similarity distribution is stable during fine-tuning for all potential images and texts. Given a teacher model $\overline{f}$, distillation loss can be applied to penalize changes from the original distribution:
\begin{equation}
    \mathcal{L}_{\text{dist\_img}}=\text{CE}(\bm{p},\bm{\overline{p}})=-\sum_{j=1}^m\bm{p}_j\cdot\log\bm{\overline{p}}_j, \label{eq:distimg}
\end{equation}
where $\overline{\bm{p}}$ is the distribution given by the teacher model.

Although the distillation form has been widely used in previous methods~\cite{hinton2015distilling,li2017learning,rebuffi2017icarl}, they are applied to continual learning from scratch. We investigate different components of the distillation loss for enhancing pre-trained vision-language models.

Three components are discussed in this paper in detail: the data source, the teacher model, and the loss design. \cref{subsec:main_properties} shows the performance for the different choices. First, for the data source, LwF~\cite{li2017learning} uses data from the current task, while iCarl~\cite{rebuffi2017icarl} carefully selects data from previous tasks. However, data from downstream tasks span a small subspace and are not good enough to preserve the whole feature space. An ideal choice is the pre-training dataset. However, the CLIP pre-training dataset is private, and the size is too large to load. To solve this challenge, we introduce the reference dataset. A reference dataset is a publicly available image dataset with enough semantics. Enough semantics can be seen as random sampling in the whole feature space. The texts can be related class names, unrelated sentences, or even random tokens. This is because text semantics are easier to sample, and we need not ground truth to calculate the distance between one image to sufficient text samples spread among feature space.

For the teacher model, \cite{li2017learning,rebuffi2017icarl} use the model after learning task i-1 and before learning task i as the teacher model. During the continual training of the vision-language model, the feature space deviates gradually from the initial model. Using fine-tuned models as teacher models enlarges this change. In contrast, we find using the pre-trained model as a teacher model not only preserves the zero-shot transfer ability but also takes advantage of well-learned feature space for better downstream performance.

Finally, previous distillation loss is applied on traditional backbones, where a classification head gives the probability for different labels. For the vision-language model, the probability is calculated based on the relative distance between images and texts. Thus, in addition to \cref{eq:distimg}, we impose regularization $\mathcal{L}_{\text{dist\_txt}}$ on the distances from a text to a batch of images. The whole framework is shown in \cref{fig:framework} (a) with the following training loss: \begin{equation}
    \mathcal{L} = \mathcal{L}_{\text{ce}}+\lambda\cdot (\mathcal{L}_{\text{lwf\_img}} + \mathcal{L}_{\text{lwf\_txt}}).
\end{equation}

\subsection{Weight Ensemble in Parameter Space}

\begin{figure}[t]
    \centering
    \includegraphics[width=0.8\columnwidth]{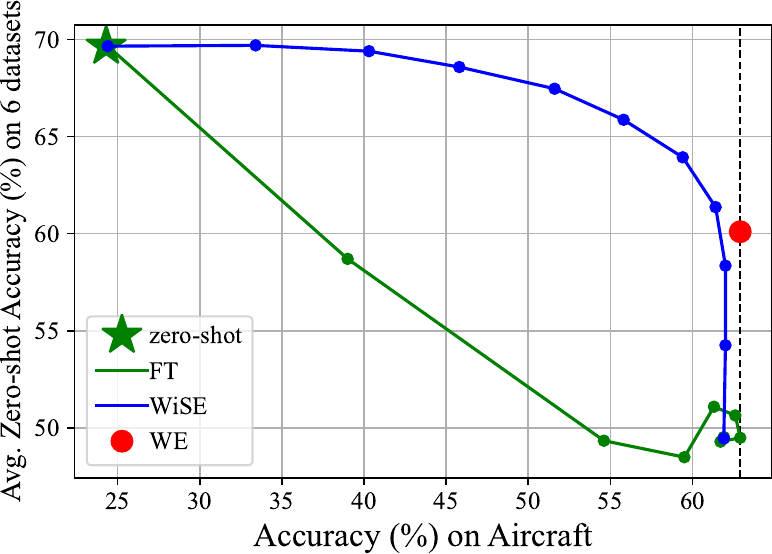}
    \caption{Models during training contain different tradeoffs between zero-shot and new task performance. Points for FT are sampled every 100 iterations, and the ones for WiSE-FT means different $\alpha$ choices. WE ensembles models during training and achieve better performance.}
    \label{fig:weightavg}
\end{figure}

Machine learning models integrate learned knowledge in their parameters. To mitigate the forgetting problem, a series of works~\cite{kirkpatrick2017overcoming,zenke2017continual,aljundi2018memory} impose regularization losses on the changes of parameters. Weight consolidation (WC)~\cite{kirkpatrick2017overcoming} imposes the following loss: 
\begin{equation}
    \mathcal{L}_{\text{WC}}=\sum_i(\theta_i - \overline{\theta}_i)^2.
\end{equation}
where $\theta$ is the parameters of the current model, and $\overline{\theta}$ is the reference ones. Although this regularization prevents forgetting, it hinders learning new tasks in a challenging CL setting.

Apart from regularization losses, another method in parameter space is to ensemble different model weights. Model soup~\cite{wortsman2022model} averages weights of multiple fine-tuned models to improve the model's robustness but introduces additional training costs. WiSE-FT~\cite{wortsman2022robust} propose a weighted average between fine-tuned model and the original model to improve the out-of-distribution prediction robustness:
\begin{equation}
    f(x; (1-\alpha)\cdot \theta_0 + \alpha\cdot\theta_1),
\end{equation}
where $\theta_0$ is the original model and $\theta_1$ is the fine-tuned one. However, this method is hyper-parameter-sensitive where different $\alpha$ gives different tradeoffs between zero-shot transfer ability and downstream task performance (blue line in \cref{fig:weightavg}.

Inspired by this, we extend the weighted average to the CL setting. The motivation for the weighted average is to prevent fine-tuning from losing too much knowledge in the original model. As training goes by (green line in \cref{fig:weightavg}), the model performs better on new tasks while losing accuracy on previous ones. The models among training compose a sequence of different learning-forgetting tradeoffs. Instead of interpolating only between the initial and the final model, our method weight ensemble (WE) averages the weights in the sequence during the training time:
\begin{equation}
    \hat{\theta}_t = \begin{cases}
        \theta_0 & t=0 \\
        \frac{1}{t+1}{\theta}_{t} + \frac{t}{t+1}\cdot\hat{\theta}_{t-1} & \text{every I iterations}
    \end{cases}.
\end{equation}
where model weight sampling happens every $I$ iteration. 
The method is related to Stochastic Weight Averaging (SWA)~\cite{izmailov2018averaging}, but we do not use a modified learning rate schedule here because instead of getting better generalization ability, WE aim to give an improved learning-forgetting tradeoff. 
As shown in \cref{fig:weightavg}, WE achieves better performance on downstream tasks than WiSE-FT. In addition, while WiSE-FT is sensitive to different values of $\alpha$, our method is much more robust under different hyper-parameter ($I$) choices. 
\section{Multi-domain Task Incremental Learning}
\label{sec:setting}

\begin{figure}[t]
    \centering
    \includegraphics[width=\columnwidth]{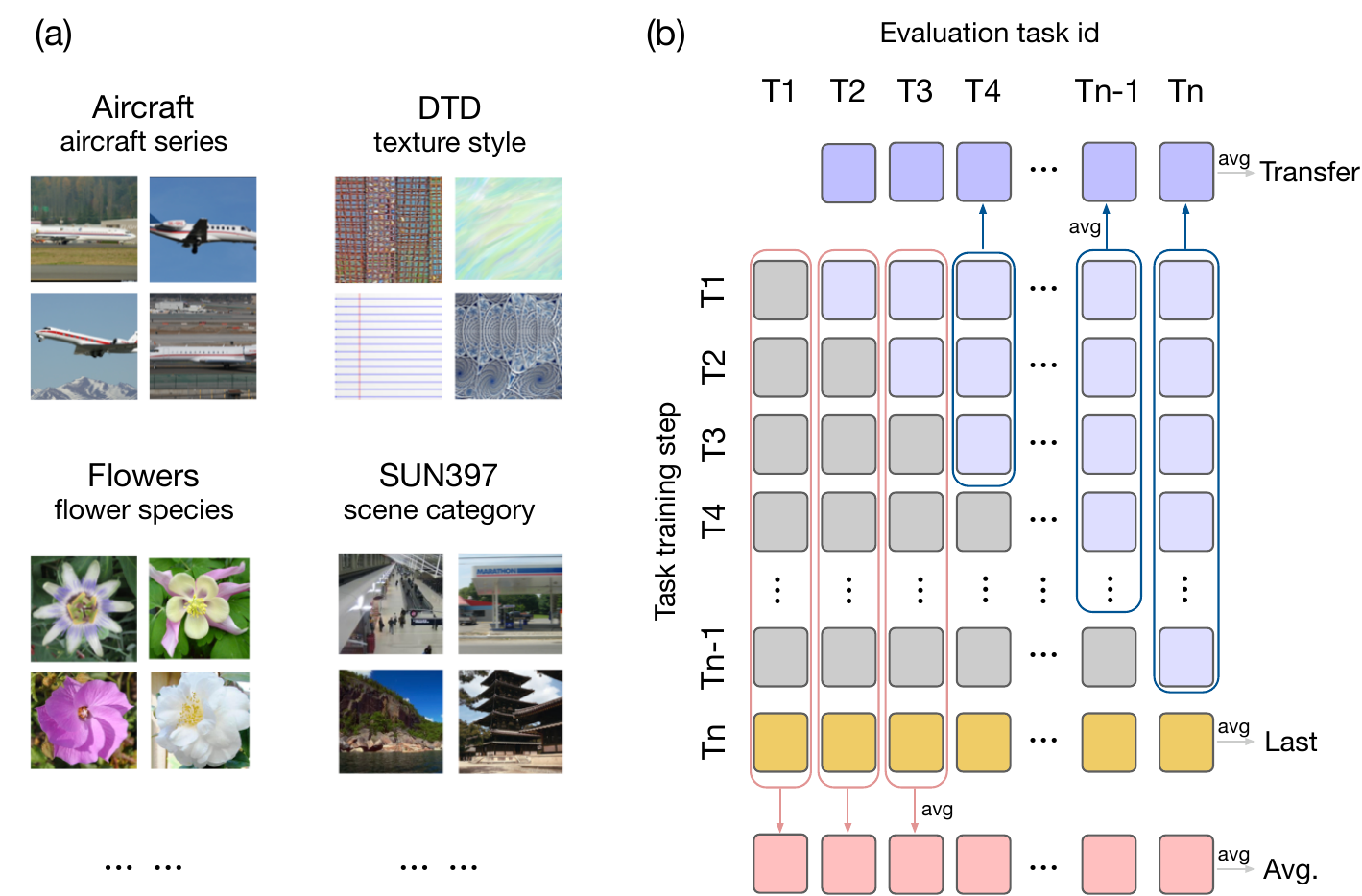}
    \caption{Fig.\textbf{(a)}: examples of tasks from different domains in MTIL benchmark. Fig.\textbf{(b)}: illustration of calculating metrics Transfer, Avg. and Last during continual learning.}
    \label{fig:benchmark}
\end{figure}

\begin{table*}[t]
\centering
\caption{
\textbf{Ablation experiments}. 
Default settings are marked in \colorbox{baselinecolor}{gray}, which uses image and text distillation loss with the initial CLIP model on 100k ImageNet images and texts generated from ImageNet classes with a simple template.
}
\subfloat[
    \textbf{Continual learning loss}.
    \label{tab:loss_type}
]{
    \centering
    \begin{minipage}{0.29\linewidth}{
        \begin{center}
            \tablestyle{1pt}{1.05}
            \begin{tabular}{y{50}x{30}x{30}x{30}}
                loss & Transfer & Avg. & Last \\
                \shline
                CE only & 44.6 & 55.9 & 77.3 \\
                Feat. Dist. & 47.6 & 58.7 & 77.1 \\
                Image-only & 56.5 & 68.9 & 82.1 \\
                Text-only & 56.7 & 69.0 & 82.6 \\
                Both & \baseline{\textbf{56.8}} & \baseline{\textbf{69.2}} & \baseline{\textbf{83.0}}
        \end{tabular}
        \end{center}
    }\end{minipage}
}
\hspace{2em}
\subfloat[
\textbf{Data sources for replay}.
\label{tab:image_source}
]{
\centering
\begin{minipage}{0.29\linewidth}{\begin{center}
\tablestyle{4pt}{1.05}
\begin{tabular}{y{30}x{30}x{30}x{30}}
source & Transfer & Avg. & Last \\
\shline
    current & 56.7 & 66.5 & 80.2 \\
    ImageNet & \baseline{56.8} & \baseline{\textbf{69.2}} & \baseline{\textbf{83.0}} \\
    CC & \textbf{57.2} & 68.5 & 80.9 \\
    CIFAR100 & 55.2 & 65.9 & 80.7 \\
    Flowers & 54.7 & 66.0 & 80.8 \\
\end{tabular}
\end{center}}\end{minipage}
}
\hspace{2em}
\subfloat[
    \textbf{Text sources for replay}.
    \label{tab:text_source}
]{
    \centering
    \begin{minipage}{0.29\linewidth}{
        \begin{center}
            \tablestyle{1pt}{1.05}
            \begin{tabular}{y{54}x{30}x{30}x{30}}
                source & Transfer & Avg. & Last \\
                \shline
                current & 51.8 & 64.9 & 82.0 \\
                prev. all & 54.0 & 70.2 & 83.7 \\
                1k classes (IN) & \baseline{56.8} & \baseline{69.2} & \baseline{83.0} \\
                13k Sent. (CC) & \textbf{58.9} & \textbf{70.5} & \textbf{84.0} \\
                1k Rand. Sent. & 58.7 & 70.2 & 83.8 \\
            \end{tabular}
        \end{center}}\end{minipage}
}
\\
\centering
\subfloat[
\textbf{Teacher model}.
\label{tab:ref_model}
]{
\centering
\begin{minipage}{0.29\linewidth}{\begin{center}
\tablestyle{4pt}{1.05}
\begin{tabular}{x{30}x{30}x{30}x{30}}
source & Transfer & Avg. & Last \\
\shline
    Initial & \baseline{\textbf{56.8}} & \baseline{\textbf{69.2}} & \baseline{\textbf{83.0}} \\
    $n-1$ & 53.9 & 66.6 & 80.7 \\
    WiSE(0.5) & 56.4 & 68.9 & 82.9 \\
    WiSE(0.8) & 56.2 & 67.8 & 81.3 
\end{tabular}
\end{center}}\end{minipage}
}
\hspace{2em}
\subfloat[
\textbf{\# images for replay}.
\label{tab:replay_num}
]{
\begin{minipage}{0.29\linewidth}{\begin{center}
\tablestyle{4pt}{1.05}
\begin{tabular}{x{30}x{30}x{30}x{30}}
\#image & Transfer & Avg. & Last \\
\shline
1M & \textbf{58.7} & \textbf{70.1} & \textbf{83.2} \\
100k & \baseline{56.8} & \baseline{69.2} & \baseline{83.0} \\
10k & 57.8 & 68.7 & 81.2 \\
1k & 56.3 & 67.6 & 80.8 \\
\end{tabular}
\end{center}}\end{minipage}
}
\hspace{2em}
\subfloat[
\textbf{\# image classes for replay}.
\label{tab:replay_classes}
]{
\begin{minipage}{0.29\linewidth}{\begin{center}
\tablestyle{4pt}{1.05}
\begin{tabular}{x{30}x{30}x{30}x{30}}
\#class & Transfer & Avg. & Last \\
\shline
1000 & \baseline{\textbf{56.8}} & \baseline{\textbf{67.6}} & \baseline{\textbf{83.0}} \\
100 & 56.7 & 67.3 & 82.3 \\
10 & 53.8 & 66.4 & 81.0 \\
1 & 53.1 & 65.5 & 80.5 
\end{tabular}
\end{center}}\end{minipage}
}
\label{tab:ablations}
\end{table*}

\paragraph{Conventional Continual Learning Benchmark.} A benchmark consisting of several tasks is needed to evaluate different methods for continual learning. Most previous benchmarks are built by separating classes in a single dataset, such as MNIST~\cite{van2019three}, CIFAR100~\cite{douillard2022dytox}, TinyImageNet~\cite{yan2021dynamically}, and ImageNet~\cite{yan2021dynamically,zhu2021prototype}. We also evaluate our method with traditional benchmarks. In CIFAR100~\cite{Krizhevsky09learningmultiple}, classes are separated into groups to build tasks. Suppose the dataset has $m$ classes, a $k$-step setting means we learn $\nicefrac{m}{k}$ classes in each new task. The CIFAR100 contains 100 classes, and the setting of 10, 20, and 50 steps are visited. For TinyImageNet with $m=200$, the first step learns 100 classes, and the rest is learned with 5, 10, and 20 steps. As for ImageNet-100, we have two settings: ImageNet-100-B0, which includes the same amount of classes for each step, and ImageNet-100-B50, which has 50 classes for the first step, and the remaining 50 classes are observed progressively over the next 10 stages. For ImageNet~\cite{deng2009imagenet}, we investigate a 10-step setting, which learns 100 new classes per task.

\paragraph{MTIL Benchmark.} Different classes from one dataset share the common image source and a similar style~\cite{recht2019imagenet,hendrycks2021many}. Thus, we propose Multi-domain Task Incremental Learning (MTIL), a cross-domain version of task incremental learning. Different tasks are collected from different domains, requiring different domain knowledge for humans to achieve high accuracy. Our MTIL benchmark consists of 11 tasks (detailed in supplementary materials), as some of the tasks illustrated in \cref{fig:benchmark} (a). The MTIL benchmark is very challenging with a total number of 1,201 classes. Two orders are used for the evaluation; the first one is alphabet order (Order-I): Aircraft, Caltech101, CIFAR100, DTD, EuroSAT, Flowers, Food, MNIST, OxfordPet, StanfordCars, SUN397. And the second one is a random order (Order-II): StanfordCars, Food, MNIST, OxfordPet, Flowers, SUN397, Aircraft, Caltech101, DTD, EuroSAT, CIFAR100. Experiments are done in Order I by default.

\begin{table}[t]
\centering
\caption{Ablation study of different components for ZSCL.}
\label{tab:pablate}
\resizebox{\columnwidth}{!}{%
\begin{tabular}{@{}lcc|cc|cc@{}}
\toprule
Method & Transfer & $\Delta$ & Avg. & $\Delta$ & Last & $\Delta$  \\ \midrule
CLIP \texttt{ViT-B/16@224px}  & & & & & & \\
\quad Zero-shot  & \baseline{69.4} &  \baseline{0.0} & \baseline{65.3} & \baseline{0.0} & \baseline{65.3} & \baseline{0.0} \\
\quad Continual Learning & 44.6 & \red{-24.8} & 55.9 & \red{-9.4} & 77.3 & \gre{+12.0} \\
\quad + Distillation & 58.9 & \red{-10.5} & 70.5 & \gre{+5.2} & 83.8 & \gre{+18.5}  \\
\quad \quad + WiSE-FT (best $\alpha$) & 61.7 & \red{-7.7} & 71.6 & \gre{+6.3} & 83.3 & \gre{+18.0}\\
\quad \quad + WE (ZSCL$^*$) & 62.2 & \red{-7.2} & 72.6 & \gre{+7.3} & \textbf{84.5} & \gre{\textbf{+19.2}} \\
\quad \quad + WC & 67.6 & \red{-1.8} & 74.5 & \gre{+9.2} & 83.2 & \gre{+17.9}\\
\quad \quad \quad + WiSE-FT & 67.7 & \red{-1.7} & 74.2 & \gre{+8.9} & 81.9 & \gre{+16.6} \\
\quad \quad \quad + WE (ZSCL) & \textbf{68.1} & \red{\textbf{-1.3}} & \textbf{75.4} & \gre{\textbf{+10.1}} & 83.6 & \gre{+18.3} \\
\bottomrule
\end{tabular}%
}
\end{table}

\paragraph{Evaluation Metrics.} The metrics of MTIL are illustrated in \cref{fig:benchmark}(b), where rows represent training steps and a column shows performance for one dataset.
For conventional continual learning, only scores under the diagonal are meaningful, since they cannot give zero-shot predictions on unseen tasks. In contrast, the zeros-hot transfer ability enables a vision-language model to provide predictions for all datasets. The average accuracy on all datasets among all timestamps is the ``Avg'' metric. The ``Last'' metric is the average performance of all tasks after CL. The ``Transfer'' metric is the average task performance in the upper-right triangle of the matrix. Every task's performance is first averaged to equal the weight of each dataset. It measures to what extent the zero-shot transfer ability is preserved. Before learning task $i$, tasks not earlier than $i$ are not fine-tuned. Thus, their performance is an indicator of zero-shot transfer ability. 
\section{Experiments}
\label{sec:exp}

\begin{figure*}[t]
    \centering
    \includegraphics[width=\textwidth,trim={0 10pt 0 0pt},clip]{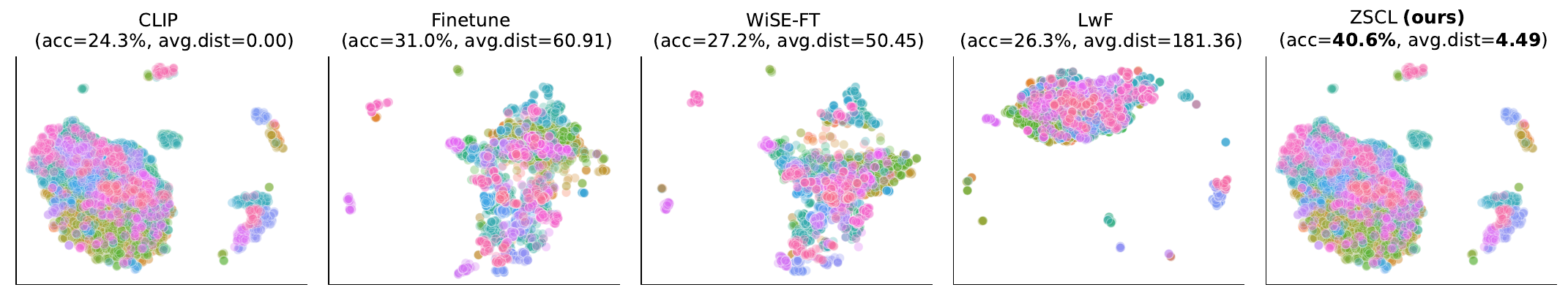}
    \caption{t-SNE on five models' outputs together of Aircraft datasets after MTCL training: only our model maintains a similar feature distribution to the original CLIP ones with minor shift, while the rest significantly distort the feature space.}
    \label{fig:tsne}
\end{figure*}

\paragraph{Implementation.} We use CLIP~\cite{radford2021learning} model with image encoder ViT-B/16~\cite{dosovitskiy2020image}. We conduct training with AdamW~\cite{loshchilov2017decoupled} optimizer and use a label smoothing~\cite{muller2019does} technique for a better baseline result. For multi-domain task continual learning, we train 1K iterations for each task, while for class-incremental learning, we follow the same evaluation setting in~\cite{douillard2022dytox}. More implementation details can be found in the supplementary material.

\subsection{Main Properties}
\label{subsec:main_properties}

We ablate our method in feature-space in \cref{tab:ablations}, and different choices for parameter-space regularization in \cref{tab:pablate}. Several interesting characteristics are noted.

\paragraph{Continual learning loss.} In \cref{tab:loss_type}, several types of loss for feature space are tested. The Feature Distance penalizing absolute distances achieves a low accuracy. Distillation loss on probability distribution regularizes relative distance in the feature space. With image-only or text-only distillation, the Transfer, Avg., and Last accuracy all improve. A further boost in performance occurs with both of the distillation losses.

\begin{table}[t]
\centering
\caption{Comparison of different methods on MTIL in Order I. 
}
\label{tab:result}
\resizebox{\columnwidth}{!}{%
\begin{tabular}{@{}lcc|cc|cc@{}}
\toprule
Method & Transfer & $\Delta$ & Avg. & $\Delta$ & Last & $\Delta$  \\ \midrule
CLIP \texttt{ViT-B/16@224px}  & & & & & &  \\
\quad Zero-shot  & \baseline{69.4} &  \baseline{0.0} & \baseline{65.3} & \baseline{0.0} & \baseline{65.3} & \baseline{0.0} \\
\quad Continual Learning & 44.6 & \red{-24.8} & 55.9 & \red{-9.4} & 77.3 & \gre{+12.0} \\
\quad \quad LwF~\cite{li2017learning} & 56.9 & \red{-12.5} & 64.7 & \red{-0.6} & 74.6 & \gre{+9.0} \\
\quad \quad iCaRL~\cite{rebuffi2017icarl} & 50.4 & \red{-19.0} & 65.7 & \gre{+0.4} & 80.1 & \gre{+14.8} \\
\quad \quad LwF-VR~\cite{ding2022don} & 57.2 & \red{-12.2} & 65.1 & \red{-0.2} & 76.6 & \gre{+11.3}  \\
\quad \quad WiSE-FT~\cite{wortsman2022robust} & 52.3 & \red{-17.1} & 60.7 & \red{-4.6} & 77.7 & \gre{+12.4} \\
\midrule
\quad \quad ZSCL$^*$ (Ours) & 62.2 & \red{-7.2} & 72.6 & \gre{+7.3} & \textbf{84.5} & \gre{\textbf{+19.2}} \\
\quad \quad ZSCL (Ours) & \textbf{68.1} & \red{\textbf{-1.3}} & \textbf{75.4} & \gre{\textbf{+10.1}} & 83.6 & \gre{+18.3} \\
\bottomrule
\end{tabular}%
}
\end{table}

\paragraph{Data source for replay.} \cref{tab:text_source,tab:image_source,tab:replay_num,tab:replay_classes} seek the standard for a good reference dataset. As shown in \cref{tab:image_source}, distillation on images of current tasks achieves a good Transfer score. However, it hinders the learning on new tasks and results in a low Avg. and Last score. Images in a specific domain (\eg, Flowers) are also not good choices. General images in ImageNet and Conceptual Caption (CC) datasets are examples of good reference datasets. These images are easily available by a web crawler~\cite{thomee2016yfcc100m}. Text with more semantics can improve performance (shown in \cref{tab:text_source}). When using sentences from the Conceptual Caption dataset~\cite{sharma2018conceptual}, or even sentences randomly generated from the CLIP vocabulary, there is no ground truth target among the texts for the image from the ImageNet dataset. However, they all achieve an improvement due to more semantics. The reference image dataset does not need to be labeled, matched with the text, or seen by the CLIP model. 

Enough semantics in the reference image dataset boosts the distillation performance. In \cref{tab:replay_num,tab:replay_classes}, a smaller number of images and classes all lead to a degradation in the performance. Fewer classes of images in the reference dataset have a worse impact on the performance compared with the image numbers. To keep a reasonable memory buffer, we randomly sample 100k images from ImageNet for MTIL and use texts from CC dataset. For class-incremental learning, conceptual caption dataset's validation set (28k images) is used to avoid information leakage.

\paragraph{Teacher model.} \cref{tab:ref_model} shows the performance of distillation loss with different teacher models. Unlike conventional continual learning, the teacher model should not be the one trained on the previous task; otherwise, the deviation from the initial CLIP in the previous task may be amplified. In contrast, with the initial CLIP as the teacher model, not only is the zero-shot performance improved but the Mean and Last scores are also boosted, indicating that preserving high-quality feature space facilitates continual learning.


\begin{table}[t]
\centering
\caption{Comparison of different methods on MTIL in Order II.}
\label{tab:result_o2}
\resizebox{\columnwidth}{!}{%
\begin{tabular}{@{}lcc|cc|cc@{}}
\toprule
Method & Transfer & $\Delta$ & Avg. & $\Delta$ & Last & $\Delta$  \\ \midrule
CLIP \texttt{ViT-B/16@224px} & & & &  \\
\quad Zero-shot & \baseline{65.4} & \baseline{0.0} & \baseline{65.3} & \baseline{0.0} & \baseline{65.3} & \baseline{0.0} \\
\quad Continual Learning & 46.6 & \red{-18.8} & 56.2 & \red{-9.1} & 67.4 & \gre{+2.1} \\
\quad \quad LwF~\cite{li2017learning} & 53.2 & \red{-12.2} & 62.2 & \red{-5.2} & 71.9 & \gre{+6.6} \\
\quad \quad iCaRL~\cite{rebuffi2017icarl} & 50.9 & \red{-14.5} & 56.9 & \red{-8.4} & 71.6 & \gre{+6.3} \\
\quad \quad LwF-VR~\cite{ding2022don} & 53.1 & \red{-12.3} & 60.6 & \red{-7.4} & 68.3 & \gre{+0.9} \\
\quad \quad WiSE-FT~\cite{wortsman2022robust} & 51.0 & \red{-14.4} & 61.5 & \red{-5.9} & 72.2 & \gre{+6.9} \\
\midrule
\quad \quad ZSCL$^*$ & 59.8 & \red{-5.6} & 71.8 & \gre{+6.5} & 83.3 & \gre{+18.0} \\
\quad \quad ZSCL & \textbf{64.2} & \red{\textbf{-1.2}} & \textbf{74.5} & \gre{\textbf{+9.2}} & \textbf{83.4} & \gre{\textbf{+18.1}} \\
\bottomrule
\end{tabular}%
}
\end{table}

\begin{table*}[t]
\centering
\caption{Transfer, Avg., and Last scores (\%) of different continue training methods on MTIL benchmark.}
\label{tab:finalacc}
{
\footnotesize
\begin{tabular}{y{120}*{11}{x{18}}}
\toprule
Method & \rot{Aircraft} & \rot{Caltech101} & \rot{CIFAR100} & \rot{DTD} & \rot{EuroSAT} & \rot{Flowers} & \rot{Food} & \rot{MNIST} & \rot{OxfordPet} & \rot{Cars} & \rot{SUN397} \\ \midrule

CLIP \texttt{ViT-B/16@224px} &  &  &  &  &  &  &  &  &  &  & \\
\quad Zero-shot & 24.3 & 88.4 & 68.2 & 44.6 & 54.9 & 71.0 & 88.5 & 59.4 & 89.0 & 64.7 & 65.2 \\
\quad Fine-tune & 62.0 & 95.1 & 89.6 & 79.5 & 98.9 & 97.5 & 92.7 & 99.6 & 94.7 & 89.6 & 81.8 \\ \midrule
\midrule

\quad \textbf{Transfer} \\
\quad \quad Continual-FT & &  67.1 & 46.0 & 32.1 & 35.6 & 35.0 & 57.7 & 44.1 & 60.8 & 20.5 & 46.6 \\
\quad \quad LwF~\cite{li2017learning} & & 74.5 & 56.9 & 39.1 & 51.1 & 52.6 & 72.8 & 60.6 & 75.1 & 30.3 & 55.9 \\
\quad \quad iCaRL~\cite{rebuffi2017icarl} & & 56.6 & 44.6 & 32.7 & 39.3 & 46.6 & 68.0 & 46.0 & 77.4 & 31.9 & 60.5\\
\quad \quad LwF-VR~\cite{ding2022don} & & 77.1 & 61.0 & 40.5 & 45.3 & 54.4 & 74.6 & 47.9 & 76.7 & 36.3 & 58.6 \\
\quad \quad WiSE-FT~\cite{wortsman2022robust} & & 73.5 & 55.6 & 35.6 & 41.5 & 47.0 & 68.3 & 53.9 & 69.3 & 26.8 & 51.9  \\
\quad \quad Dist. only & & 80.1 & 62.2 & 40.2 & 39.9 & 58.1 & 80.8 & 53.4 & 74.6 & 38.1 & 61.9 \\ 
\quad \quad ZSCL$^*$ & & 78.3 & 64.0 & 42.9 & 45.2 & 63.5 & 84.2 & 56.1 & 78.9 & 44.1 & 64.3 \\
\quad \quad ZSCL & & \textbf{86.0} & \textbf{67.4} & \textbf{45.4} & \textbf{50.4} & \textbf{69.1} & \textbf{87.6} & \textbf{61.8} & \textbf{86.8} & \textbf{60.1} & \textbf{66.8} \\ \midrule

\quad \textbf{Avg.} \\
\quad \quad Continual-FT & 25.5 & 81.5 & 59.1 & 53.2 & 64.7 & 51.8 & 63.2 & 64.3 & 69.7 & 31.8 & 49.7 \\
\quad \quad LwF~\cite{li2017learning} & 36.3 & 86.9 & 72.0 & 59.0 & 73.7 & 60.0 & 73.6 & 74.8 & 80.0 & 37.3 & 58.1 \\
\quad \quad iCaRL~\cite{rebuffi2017icarl} & 35.5 & 89.2 & 72.2 & 60.6 & 68.8 & 70.0 & 78.2 & 62.3 & 81.8 & 41.2 & 62.5 \\
\quad \quad LwF-VR~\cite{ding2022don} & 29.6 & 87.7 & 74.4 & 59.5 & 72.4 & 63.6 & 77.0 & 66.7 & 81.2 & 43.7 & 60.7 \\
\quad \quad WiSE-FT~\cite{wortsman2022robust} & 26.7 & 86.5 & 64.3 & 57.1 & 65.7 & 58.7 & 71.1 & 70.5 & 75.8 & 36.9 & 54.6 \\
\quad \quad Dist. only & 48.1 & 90.6 & 79.8 & 63.2 & 75.6 & 72.5 & 84.7 & 70.2 & 79.8 & 46.9 & 63.7 \\ 
\quad \quad ZSCL$^*$ & \textbf{50.7} & 90.9 & 79.8 & 63.8 & 76.6 & 77.3 & 87.0 & 71.9 & 83.0 & 52.0 & 65.9 \\
\quad \quad ZSCL & 45.1 & \textbf{92.0} & \textbf{80.1} & \textbf{64.3} & \textbf{79.5} & \textbf{81.6} & \textbf{89.6} & \textbf{75.2} & \textbf{88.9} & \textbf{64.7} & \textbf{68.0} \\ \midrule

\quad \textbf{Last} \\
\quad \quad Continual-FT & 31.0 & 89.3 & 65.8 & 67.3 & 88.9 & 71.1 & 85.6 & 99.6 & 92.9 & 77.3 & 81.1 \\
\quad \quad LwF~\cite{li2017learning} & 26.3 & 87.5 & 71.9 & 66.6 & 79.9 & 66.9 & 83.8 & 99.6 & 92.1 & 66.1 & 80.4 \\
\quad \quad iCaRL~\cite{rebuffi2017icarl} & 35.8 & \textbf{93.0} & 77.0 & 70.2 & 83.3 & 88.5 & 90.4 & 86.7 & 93.2 & 81.2 & \textbf{81.9} \\
\quad \quad LwF-VR~\cite{ding2022don} & 20.5 & 89.8 & 72.3 & 67.6 & 85.5 & 73.8 & 85.7 & 99.6 & 93.1 & 73.3 & 80.9 \\
\quad \quad WiSE-FT~\cite{wortsman2022robust} & 27.2 & 90.8 & 68.0 & 68.9 & 86.9 & 74.0 & 87.6 & 99.6 & 92.6 & 77.8 & 81.3 \\
\quad \quad Dist. only & 43.3 & 91.9 & \textbf{81.3} & \textbf{72.4} & \textbf{95.1} & 90.5 & 90.4 & \textbf{99.7} & 92.5 & 85.1 & 81.8 \\ 
\quad \quad ZSCL$^*$ & \textbf{46.0} & 92.3 & 81.2 & \textbf{72.4} & 93.0 & \textbf{92.1} & 90.8 & 99.6 & 93.3 & \textbf{86.6} & 81.7 \\
\quad \quad ZSCL & 40.6 & 92.2 & \textbf{81.3} & 70.5 & 94.8 & 90.5 & \textbf{91.9} & 98.7 & \textbf{93.9} & 85.3 & 80.2 \\
\bottomrule
\end{tabular}%
}
\end{table*}

\paragraph{Parameter-space regularization.} In~\cref{tab:pablate}, we experiment with three different parameter-space regularizations. 
We experiment with two variants of WiSE-FT~\cite{wortsman2022robust}: the weighted average between the current model with the initial CLIP or the one at the previous task. The result shows the latter one is a better choice because keeping weight averaging with initial CLIP loses the newly learned knowledge. 
We experiment with different $\alpha$ choices for WiSE, and the best result is reported. While distillation loss improves the whole performance, the parameter-space regularization further protects the zero-shot transfer ability with a higher Transfer. Among the three parameter-space regularizations, only WE achieve a better Last score.
WC greatly improves the Transfer scores with a lower Last score. A combination of the weight consolidation loss with weight ensemble achieves a better tradeoff between Transfer and Last value. While ZSCL$^*$, a variant without the WC loss, obtains the highest Last score, the ZSCL with WC loss outperforms it with 2.8\% Avg. and 5.9\% Transfer scores.

\subsection{Multi-domain Task Incremental Learning}
\label{subsec:mtcl}

\cref{tab:result} displays the performance of different methods on the MTIL benchmark, and \cref{tab:finalacc} presents the detailed Transfer, Avg, and Last metrics on each dataset. Zero-shot denotes the zero-shot prediction performance of the initial CLIP model, and Fine-tune means the direct fine-tuning accuracy on each dataset, which can be seen as an upper-bound where no forgetting phenomenon happens. Continual learning uses cross-entropy loss to learn each dataset sequentially, where there is a significant forgetting issue on both zero-shot predictions (Transfer drops by 24.8\%) and newly learned knowledge (Last drops by 9.4\%). Previous methods improve the Last performance slightly and cannot maintain a high zero-shot prediction performance. Without WC, ZSCL$^*$ achieves the best Last scores, outperforming the previous best one by 4.4\%. Our method ZSCL improves 9.1\% on Transfer accuracy, with only 1.3\% drops compared with the initial CLIP model, and achieves a 10.1\% gain in the Avg. accuracy.

Figure~\cref{fig:tsne} provides a visualization of the feature space on Aircraft of the original CLIP and four methods trained at the end of MTCL (t-SNE conducted only once on all features collected). This shows our method is capable of maintaining the pretrained feature distribution with a small averaged feature distance and thus preserving zero-shot performance.

The result of the MTIL method in Order-II is presented in \cref{tab:result_o2}. Our method surpasses previous methods in another order setting of the MTIL benchmark. A similar conclusion holds from the results of MTIL Order-II compared with MTIL Order-I. Our method ZSCL outperforms others by 9.2\% on the Avg. score and 18.1\% on the Last score with only a 1.2\% performance loss on the Transfer score. This shows our approach can work for different orders of the multi-domain task incremental learning. In addition, compared with Order-I, previous methods achieve a much lower Last score (e.g., for Continual-Learning, Order-I has 77.3\%, while Order-II has 65.3\%). With ZSCL, the Last score is similar (83.6\% compared with 83.4\%). This shows our method is more robust towards different training orders.

\begin{table}[t]
  \begin{center}
    \caption{Comparison of state-of-the-art CL methods on CIFAR100 benchmark in class-incremental setting.}
    \label{tab:cifar100_class_incre}
    \resizebox{\columnwidth}{!}{%
    \begin{tabular}{y{55}{c}c|cc|cc} 
    \toprule
         & \multicolumn{2}{c}{{ 10 steps}} & \multicolumn{2}{c}{{20 steps}} & \multicolumn{2}{c}{50 steps}  \\
    \textbf{Methods} & \textbf{Avg} & \textbf{Last} & \textbf{Avg} & \textbf{Last} & \textbf{Avg} & \textbf{Last} \\
    \midrule
    UCIR~\cite{hou2019learning} & 58.66 & 43.39 & 58.17 & 40.63 & 56.86 & 37.09 \\
    BiC~\cite{wu2019large} & 68.80 & 53.54 & 66.48 & 47.02 & 62.09 & 41.04 \\
    RPSNet~\cite{rajasegaran2019adaptive}  & 68.60 & 57.05 & -  & -  & -  & -  \\
    PODNet~\cite{douillard2020podnet} & 58.03 & 41.05 & 53.97 & 35.02 & 51.19 & 32.99 \\
    DER~\cite{yan2021dynamically} & \underline{74.64} & 64.35 & 73.98 & 62.55 & 72.05 & 59.76 \\
    DyTox+~\cite{douillard2022dytox} & 74.10 & 62.34 & 71.62 & 57.43 & 68.90 & 51.09 \\
    \midrule
    CLIP~\cite{radford2021learning} & 74.47 & \underline{65.92} & \underline{75.20} & \underline{65.74} & \underline{75.67} & \underline{65.94} \\
    FT & 65.46 & 53.23 & 59.69 & 43.13 & 39.23 & 18.89 \\
    LwF~\cite{li2017learning} & 65.86 & 48.04 & 60.64 & 40.56 & 47.69 & 32.90   \\
    iCaRL~\cite{rebuffi2017icarl} & 79.35 & 70.97 & 73.32 & 64.55 & 71.28 & 59.07 \\
    LwF-VR~\cite{ding2022don} & 78.81 & 70.75 & 74.54 & 63.54 & 71.02 & 59.45 \\
    ZSCL (Ours) & \textbf{82.15} & \textbf{73.65} & \textbf{80.39} & \textbf{69.58}  & \textbf{79.92} & \textbf{67.36}   \\
    \textbf{Impr} & \gre{\textbf{+7.68}} & \gre{\textbf{+7.73}} & \gre{\textbf{+5.19}} & \gre{\textbf{+3.84}} & \gre{\textbf{+3.95}} & \gre{\textbf{+1.42}} \\
    \bottomrule
    \end{tabular}}
\end{center}
\end{table}

\subsection{Class Incremental Learning}
\label{subsec:cil}

We evaluate our methods on conventional class incremental learning.
\cref{tab:cifar100_class_incre,tab:tiny_imagenet_class_incre} display results on CIFAR100 and TinyImageNet, respectively. 
We re-implement some previous methods with a CLIP backbone (after CLIP in the table), while others using a special network design cannot be easily adapted. Although zero-shot CLIP prediction achieves a good result on these benchmarks, continual learning with direct fine-tuning or LwF~\cite{li2017learning} degrades the performance greatly, especially under the setting of a large step number. This demonstrates a severe catastrophic forgetting phenomenon in fine-tuning the CLIP model. Our method consistently improves the performance of the CLIP model on both Avg. and Last scores with a large gap towards previous ones. 

\section{Limitation and Future Work}

\begin{table}
  \begin{center}
    \caption{Comparison of different methods on TinyImageNet splits in class-incremental settings with 100 base classes.}
    \label{tab:tiny_imagenet_class_incre}
    \resizebox{\columnwidth}{!}{%
    \begin{tabular}{y{55}{c}c|cc|cc} 
  \toprule
    & \multicolumn{2}{c}{{5 steps}} &  \multicolumn{2}{c}{10 steps}    &  \multicolumn{2}{c}{20 steps}  \\
    \textbf{Methods}    & \textbf{Avg}   & \textbf{Last}    & \textbf{Avg}   & \textbf{Last}    & \textbf{Avg}   & \textbf{Last}  \\
    \midrule
    EWC~\cite{kirkpatrick2017overcoming}       & 19.01 & 6.00  & 15.82 & 3.79  & 12.35 & 4.73 \\
    EEIL~\cite{castro2018end}          & 47.17 & 35.12 & 45.03 & 34.64 & 40.41 & 29.72 \\
    UCIR~\cite{hou2019learning}               & 50.30 & 39.42 & 48.58 & 37.29 & 42.84 & 30.85 \\
    MUC~\cite{liu2020more}                    & 32.23 & 19.20  & 26.67 & 15.33 & 21.89 & 10.32 \\
    PASS~\cite{zhu2021prototype}        & 49.54 & 41.64 & 47.19 & 39.27 & 42.01 & 32.93 \\
    DyTox~\cite{douillard2022dytox}           & 55.58 & 47.23 & 52.26 & 42.79 & 46.18 & 36.21 \\
    \midrule
    CLIP~\cite{radford2021learning}  & \underline{69.62} & \underline{65.30} & \underline{69.55} & \underline{65.59} & \underline{69.49} & \underline{65.30} \\ 
    FT & 61.54 &46.66  &57.05 &41.54 & 54.62 & 44.55\\
    LwF~\cite{li2017learning} &60.97 &48.77 &57.60 &44.00 &54.79 &42.26 \\
    iCaRL~\cite{rebuffi2017icarl} & 77.02 & 70.39 & 73.48 & 65.97 & 69.65 & 64.68 \\
    LwF-VR~\cite{ding2022don} & 77.56 & 70.89 & 74.12 & 67.05 & 69.94 & 63.89 \\
    ZSCL (Ours) & \textbf{80.27} & \textbf{73.57} & \textbf{78.61} & \textbf{71.62} & \textbf{77.18} & \textbf{68.30} \\
    \textbf{Impr} & \hspace{-4pt}\gre{\textbf{+10.65}} & \gre{\textbf{+8.27}} & \gre{\textbf{+9.06}} & \gre{\textbf{+6.03}} & \gre{\textbf{+7.69}} & \gre{\textbf{+3.00}} \\
    \bottomrule
    \end{tabular}}
\end{center}
\end{table}

Our proposed method has a limitation that a reference dataset is needed. A promising direction of the work is to preserve the zero-shot transfer ability without the need for an outside dataset. For example, we may generate a synthetic image dataset as the reference dataset. Methods like~\cite{smith2021always} can synthesize datasets from a network.

The deep learning community tends to build large models with a huge dataset~\cite{brown2020language,dehghani2023scaling}, including vision-language models~\cite{radford2021learning,huang2023language}. 
As re-training cost upsurges, continual learning is an efficient approach for updating these models with custom usage.

In some cases, we want to correct wrong information in the pre-training dataset or update old information with the latest one. How to conduct this task with a reference dataset is left as future work.

Lately, a noticeable trend involves the creation of multi-modality models utilizing large language models, showcasing encouraging outcomes in tasks such as Visual Question Answering (VQA). Expanding our approach to encompass the next token prediction task remains an avenue for future exploration and research.

\section{Conclusion}
\label{sec:conclusion}

In this paper, we investigate continual learning with the vision-language model. We propose a better continual learning algorithm to protect the zero-shot transfer ability in the vision-language model learned in the pre-training stage. Our algorithm mitigates the catastrophic forgetting in both feature space and parameter space. In feature space, distilling the initial model on a reference dataset significantly boosts the model's performance. In parameter space, weight ensemble among different training stages alleviates the forgetting issue. We propose a more challenging Multi-domain Task Incremental Learning (MTIL) benchmark to evaluate the continual learning methods better. On both conventional and new benchmarks, our method achieves state-of-the-art performance.

\section*{Acknowledgements}

Yang You's research group is being sponsored by the NUS startup grant (Presidential Young Professorship), Singapore MOE Tier-1 grant, ByteDance grant, ARCTIC grant, SMI grant, and Alibaba grant.

{\small
\bibliographystyle{ieee_fullname}
\bibliography{egbib}
}

\clearpage
\appendix
\noindent\textbf{\Large Appendix}
\section{Additional Benchmark Description}
\label{sec:dcd}

\cref{tab:description} displays the detailed information for different datasets in our benchmark.

\section{Additional Implementation Details}
\label{sec:addid}

Our implementation is based on PyTorch~\cite{NEURIPS2019_9015}. We use batch size 64 for the MTIL benchmark and 128 for the class-incremental learning benchmark. The learning rates are searched among $\{10^{-5},10^{-6},10^{-7}\}$. Label smoothing~\cite{muller2019does} can substitute the regularization of weight decay and achieve better performance. the label smoothing strength is searched among $\{0.1,0.2,0.3\}$. In general, for MTIL, CIFAR100, and TinyImageNet, weight decay 0 and label smoothing 0.2 are good choices. For ImageNet, weight decay 0.1 and label smoothing 0 are used. We experiment with $I\in\{1,10,100\}$ and find slight changes in performance and thus fix $I$ to 100. A large $\lambda$ hinders learning new knowledge, and $\lambda=1$ is a good choice.

\section{Additional MTIL Results}

The complete result of the MTIL benchmark with $t$ datasets is a matrix of $t\times t$. It is difficult to compare different matrices between different methods, so we summarize the performance by three indicators in the main text. Here, we show the complete matrix of ZSCL in \cref{tab:complete_res_2} and ZSCL$^*$ in \cref{tab:complete_res}.  The detailed result of the MTIL method in Order-II is presented in \cref{tab:finalacc_o2}.


\section{Additional Conventional Class Incremental Learning Results}

We re-implement previous methods based on the CLIP backbone for continual learning. For LwF-based~\cite{li2019learn,ding2022don} methods, we experiment with two choices for the teacher model, the previous one and the initial one. We find the initial one gives out better performance and report this result. For the WiSE-FT~\cite{wortsman2022robust} method, we take the average of models after learning each task. We experiment with the average between the previous and current one and the initial one and the current one. Better results are reported.

The result of class-continual learning on ImageNet benchmark is presented in \cref{tab:imagenet_class_incre}. On IN100-B10, our method outperforms others by 1.97\% for the Avg. score and 1.30\% for the Last score. On IN100-B50, our method outperforms others by 3.54\% for the Avg. score and 6.62 for the Last score.

\begin{table}[t]
\centering
\caption{Dataset description of multi-domain task incremental learning.}
\label{tab:description}
\resizebox{\columnwidth}{!}{\small
\begin{tabular}{@{}lllll@{}}
\toprule
Dataset & \# classes & \# train & \# test & Recognition Task \\ \midrule
Aircraft~\cite{maji13fine-grained} & 100 & 3334 & 3333 & aircraft series \\
Caltech101~\cite{FeiFei2004LearningGV} & 101 & 6941 & 1736 & real-life object \\
CIFAR100~\cite{Krizhevsky09learningmultiple} & 100 & 50000 & 10000 & real-life object \\
DTD~\cite{cimpoi14describing} & 47 & 1880 & 1880 & texture recognition \\
EuroSAT~\cite{helber2018novel} & 10 & 21600 & 5300 & satellite location \\
Flowers~\cite{nilsback2008automated} & 102 & 1020 & 6149 & flower species \\
Food~\cite{bossard14} & 101 & 75750 & 25250 & food type \\
MNIST~\cite{deng2012mnist} & 10 & 60000 & 10000 & digital number \\
OxfordPet~\cite{parkhi12a} & 37 & 3680 & 3669 & animal species \\
StanfordCars~\cite{KrauseStarkDengFei-Fei_3DRR2013} & 196 & 8144 & 8041 & car series \\
SUN397~\cite{xiao2010sun} & 397 & 87003 & 21751 & scene category \\ \midrule
Total & 1201 & 319352 & 97109 & \\
\bottomrule
\end{tabular}
}
\end{table}

\begin{table}[!t]
  \begin{center}
    \caption{Comparison of state-of-the-art CL methods on different ImageNet benchmarks, in class-incremental settings with 10 splits, in terms of average and last accuracy values.}
    \label{tab:imagenet_class_incre}
    \resizebox{0.9\columnwidth}{!}{
    \begin{tabular}{{l}cc|cc} 
        \toprule
         & \multicolumn{2}{c}{{ImageNet100-B10}} & 
         \multicolumn{2}{c}{ImageNet100-B50}  \\
        \textbf{Methods}    & \textbf{Avg}   & \textbf{Last}    & \textbf{Avg}   & \textbf{Last} \\ 
        \midrule
        UCIR \cite{hou2019learning}                & -     & -   & 68.09 & 57.30 \\
        TPCIL \cite{tao2020topology}               & -     & -   & 74.81 & 66.91 \\
        PODNet \cite{douillard2020podnet}          & -     & -   & 74.33 & -     \\
        DER \cite{yan2021dynamically}                   & 76.12 & 66.06 & 77.13 & 72.06 \\
        DyTox \cite{douillard2022dytox}           & 73.96 & 62.20  & -     & -     \\
        DyTox+ \cite{douillard2022dytox}          & 77.15 & 67.70 & -     & -     \\
        \midrule        
        CLIP   &84.42 &74.92 &78.86 &74.92 \\
        \quad FT &83.10  &70.72  &80.31  &72.48  \\
        \quad LwF~\cite{li2017learning} &83.35  &72.40  &80.74  &72.22    \\
        \quad iCaRL~\cite{rebuffi2017icarl} & 83.40 & 70.96 & 79.76 & 73.96 \\
        \quad LwF-VR~\cite{ding2022don} & 82.53 & 69.68 & 80.82 & 70.18 \\
        \quad ZSCL & \textbf{86.39} &  \textbf{76.22}  & \textbf{84.28} & \textbf{79.54}   \\
        \quad\textbf{Impr} &  \gre{\textbf{+1.97}} & \gre{\textbf{+1.30}} & \gre{\textbf{+3.54}} & \gre{\textbf{+6.62}} \\
        \bottomrule
    \end{tabular}}
\end{center}
\end{table}

\begin{table*}[t]
\centering
\caption{Accuracy (\%) of the ZSCL method on the MTIL benchmark with order-I. Each row represents the performance on every dataset of the model trained after the corresponding task. \colorbox{before}{Transfer}, \colorbox{avg}{Avg.}, and \colorbox{last}{Last} metrics are shown in color.}
\label{tab:complete_res}
\begin{tabular}{@{}lcccccccccccc@{}}
\toprule
& \rot{Aircraft} & \rot{Caltech101} & \rot{CIFAR100} & \rot{DTD} & \rot{EuroSAT} & \rot{Flowers} & \rot{Food} & \rot{MNIST} & \rot{OxfordPet} & \rot{Cars} & \rot{SUN397} \\ \midrule
Transfer & & 86.0 & 67.4 & 45.4 & 50.4 & 69.1 & 87.6 & 61.8 & 86.8 & 60.1 & 66.8 & 68.1\Gb \\ \midrule
Aircraft & 55.1 & 86.0 & 66.3 & 44.9 & 49.2 & 70.6 & 88.3 & 53.6 & 87.4 & 61.3 & 65.7 \\
Caltech101 & 48.9 & 94.2 & 68.6 & 44.7 & 50.4 & 67.0 & 87.6 & 55.2 & 85.0 & 61.0 & 65.9 \\
CIFAR100 & 47.1 & 93.1 & 86.2 & 46.5 & 50.1 & 68.8 & 87.7 & 63.4 & 87.6 & 60.6 & 67.4 \\
DTD & 47.0 & 93.8 & 85.0 & 76.2 & 51.7 & 69.9 & 87.8 & 65.9 & 88.4 & 61.2 & 67.2 \\
EuroSAT & 46.1 & 92.8 & 84.0 & 75.0 & 97.8 & 69.1 & 87.1 & 67.6 & 88.0 & 60.3 & 67.1 \\
Flowers & 43.8 & 92.5 & 83.2 & 73.5 & 97.2 & 96.3 & 87.1 & 63.3 & 86.5 & 58.8 & 66.9 \\
Food & 44.3 & 92.2 & 82.9 & 71.2 & 96.8 & 93.8 & 92.2 & 63.5 & 87.3 & 60.3 & 67.9 \\
MNIST & 41.9 & 91.9 & 80.5 & 67.8 & 95.3 & 89.5 & 91.9 & 99.0 & 84.4 & 58.4 & 66.5 \\
OxfortPet & 41.6 & 91.8 & 81.3 & 68.2 & 95.7 & 90.7 & 92.0 & 98.8 & 95.3 & 58.9 & 66.4 \\
Cars & 39.8 & 91.9 & 81.8 & 68.9 & 95.7 & 91.6 & 91.8 & 98.8 & 94.5 & 85.8 & 67.3 \\
SUN397 & 40.6 & 92.2 & 81.3 & 70.5 & 94.8 & 90.5 & 91.9 & 98.7 & 93.9 & 85.3 & 80.2 & 83.6\Gl \\ \midrule
Avg. & 45.1 & 92.0 & 80.1 & 64.3 & 79.5 & 81.6 & 89.6 & 75.2 & 88.9 & 64.7 & 68.0 & 75.4\Ga \\
\bottomrule
\end{tabular}%
\end{table*}

\begin{table*}[t]
\centering
\caption{Accuracy (\%) of the ZSCL$*$ method on the MTIL benchmark with order I. Each row represents the performance on every dataset of the model trained after the corresponding task. \colorbox{before}{Transfer}, \colorbox{avg}{Avg.}, and \colorbox{last}{Last} metrics are shown in color.}
\label{tab:complete_res_2}
\begin{tabular}{@{}lcccccccccccc@{}}
\toprule
& \rot{Aircraft} & \rot{Caltech101} & \rot{CIFAR100} & \rot{DTD} & \rot{EuroSAT} & \rot{Flowers} & \rot{Food} & \rot{MNIST} & \rot{OxfordPet} & \rot{Cars} & \rot{SUN397} \\ \midrule
Transfer & & 78.3 & 64.0 & 42.9 & 45.2 & 63.5 & 84.2 & 56.1 & 78.9 & 44.1 & 64.3 & 62.2\Gb \\ \midrule
Aircraft & 63.5 & 78.3 & 61.5 & 41.1 & 48.5 & 61.3 & 83.5 & 51.5 & 80.4 & 42.4 & 62.4 \\
Caltech101 & 56.1 & 93.0 & 66.6 & 43.7 & 40.3 & 64.7 & 84.7 & 57.2 & 82.4 & 47.4 & 66.0 \\
CIFAR100 & 55.5 & 92.7 & 88.7 & 44.0 & 47.6 & 62.9 & 85.0 & 58.1 & 82.9 & 48.0 & 66.8 \\
DTD & 55.3 & 92.9 & 87.8 & 77.9 & 44.5 & 66.1 & 84.8 & 57.8 & 86.2 & 49.7 & 66.6 \\
EuroSAT & 54.5 & 92.9 & 85.8 & 76.4 & 98.5 & 62.7 & 83.1 & 58.0 & 82.5 & 46.8 & 65.5 \\
Flowers & 53.5 & 92.5 & 85.5 & 76.5 & 98.1 & 97.7 & 84.2 & 55.6 & 83.2 & 47.9 & 66.6 \\
Food & 53.0 & 92.3 & 85.0 & 75.6 & 98.1 & 96.1 & 92.6 & 54.7 & 83.3 & 51.7 & 67.9 \\
MNIST & 34.2 & 89.1 & 75.1 & 57.5 & 88.6 & 66.3 & 87.1 & 99.6 & 50.4 & 24.4 & 55.0 \\
OxfortPet & 42.4 & 91.2 & 79.5 & 66.8 & 91.9 & 87.5 & 90.1 & 99.6 & 94.4 & 39.0 & 61.4 \\
Cars & 44.2 & 92.5 & 80.7 & 70.4 & 93.2 & 92.4 & 90.9 & 99.6 & 93.9 & 88.4 & 65.1 \\
SUN397 & 46.0 & 92.3 & 81.2 & 72.4 & 93.0 & 92.1 & 90.8 & 99.6 & 93.3 & 86.6 & 81.7 & 84.5\Gl \\ \midrule
Avg. & 50.7 & 90.9 & 79.8 & 63.8 & 76.6 & 77.3 & 87.0 & 71.9 & 83.0 & 52.0 & 65.9 & 72.6\Ga \\
\bottomrule
\end{tabular}%
\end{table*}

\begin{table*}[t]
\centering
\caption{Transfer, Avg., Last accuracy (\%) of different continue training methods on MTIL benchmark in Order II.}
\label{tab:finalacc_o2}
{
\footnotesize
\begin{tabular}{@{}lccccccccccc@{}}
\toprule
Method & \rot{Cars} & \rot{Food} & \rot{MNIST} & \rot{OxfordPet} & \rot{Flowers} & \rot{SUN397} & \rot{Aircraft} & \rot{Caltech101} & \rot{DTD} & \rot{EuroSAT} & \rot{CIFAR100} \\ \midrule

CLIP \texttt{ViT-B/16@224px} &  &  &  &  &  &  &  &  &  &  & \\
\quad Zero-shot & 64.7 & 88.5 & 59.4 & 89.0 & 71.0 & 65.2 & 24.3 & 88.4 & 44.6 & 54.9 & 68.2 \\
\quad Fine-tune & 89.6 & 92.7 & 94.7 & 94.7 & 97.5 & 81.8 & 62.0 & 95.1 & 79.5 & 98.9 & 89.6 \\ \midrule
\midrule

\quad \textbf{Transfer} \\
\quad \quad Continual-FT  & & 85.9 & 59.6 & 57.9 & 40.0 & 46.7 & 11.1 & 70.0 & 30.5 & 26.6 & 37.7 \\
\quad \quad LwF~\cite{li2017learning}  & & 87.8 & 58.5 & 71.9 & 46.6 & 57.3 & 12.8 & 81.4 & 34.5 & 34.5 & 46.8 \\
\quad \quad iCaRL~\cite{rebuffi2017icarl} & & 86.1 & 51.8 & 67.6 & 50.4 & 57.9 & 11.0 & 72.3 & 31.2 & 32.7 & 48.1 \\
\quad \quad LwF-VR~\cite{ding2022don}  & & 88.2 & 57.0 & 71.4 & 50.0 & 58.0 & 13.0 & 82.0 & 34.4 & 29.3 & 47.6  \\
\quad \quad WiSE-FT~\cite{wortsman2022robust}  & & 87.2 & 57.6 & 67.0 & 45.0 & 54.0 & 12.9 & 78.6 & 35.5 & 28.4 & 44.3 \\
\quad \quad ZSCL$^*$  & & \textbf{88.8} & 56.7 & 75.5 & 58.8 & 62.5 & 16.1 & 87.0 & 42.0 & 44.0 & 66.5 \\ 
\quad \quad ZSCL  & & 88.3 & \textbf{57.5} & \textbf{84.7} & \textbf{68.1} & \textbf{64.8} & \textbf{21.1} & \textbf{88.2} & \textbf{45.3} & \textbf{55.2} & \textbf{68.2} \\ \midrule

\quad \textbf{Avg.} \\
\quad \quad Continual-FT  & 42.1 & 70.5 & 92.2 & 80.1 & 54.5 & 59.1 & 19.8 & 78.3 & 41.0 & 38.1 & 42.3 \\
\quad \quad LwF~\cite{li2017learning}  & 49.0 & 77.0 & 92.1 & 85.9 & 66.5 & 67.2 & 20.9 & 84.7 & 44.6 & 45.5 & 50.5 \\
\quad \quad iCaRL~\cite{rebuffi2017icarl} & 52.0 & 75.9 & 77.4 & 74.6 & 58.4 & 59.3 & 11.7 & 79.6 & 42.1 & 43.2 & 51.7\\
\quad \quad LwF-VR  & 44.9 & 75.8 & 91.8 & 85.3 & 63.5 & 67.6 & 16.9 & 84.9 & 44.0 & 40.6 & 51.3 \\
\quad \quad WiSE-FT~\cite{wortsman2022robust}  & 52.6 & 79.3 & 91.9 & 83.9 & 63.4 & 65.2 & 23.3 & 83.7 & 45.4 & 40.0 & 48.2 \\
\quad \quad ZSCL$^*$ & 72.0 & 89.8 & \textbf{91.7} & 87.9 & 78.8 & 71.5 & \textbf{35.1} & 89.0 & 51.4 & 53.9 & 68.5 \\ 
\quad \quad ZSCL & \textbf{81.7} & \textbf{91.3} & 91.1 & \textbf{91.0} & \textbf{82.9} & \textbf{72.5} & 33.6 & \textbf{89.7} & \textbf{53.3} & \textbf{62.8} & \textbf{69.9} \\ \midrule

\quad \textbf{Last} \\
\quad \quad Continual-FT  & 24.0 & 67.3 & 99.1 & 87.4 & 44.3 & 67.0 & 29.5 & 92.3 & 61.3 & 81.0 & 88.1 \\
\quad \quad LwF~\cite{li2017learning}  & 34.6 & 69.6 & 99.3 & 88.7 & 61.1 & 72.5 & 32.5 & 88.1 & 65.6 & 90.9 & 87.9 \\
\quad \quad iCaRL~\cite{rebuffi2017icarl} & 46.0 & 81.5 & 91.3 & 82.8 & 66.5 & 72.2 & 16.3 & 91.6 & 68.1 & 83.2 & 87.8 \\
\quad \quad LwF-VR~\cite{ding2022don}  & 27.4 & 61.2 & 99.4 & 86.3 & 60.6 & 70.7 & 23.4 & 88.0 & 61.3 & 84.3 & 88.1 \\
\quad \quad WiSE~\cite{wortsman2022robust}  & 35.6 & 76.9 & 99.5 & 89.1 & 62.1 & 71.8 & 27.8 & 90.8 & 67.0 & 85.6 & 87.6 \\
\quad \quad ZSCL$^*$  & 63.5 & 89.6 & \textbf{99.2} & 92.4 & 84.5 & \textbf{78.3} & \textbf{55.2} & \textbf{92.4} & \textbf{74.6} & \textbf{97.4} & \textbf{88.6} \\ 
\quad \quad ZSCL  & \textbf{78.2} & \textbf{91.1} & 97.6 & \textbf{92.5} & \textbf{87.4} & 78.2 & 45.0 & 92.3 & 72.7 & 96.2 & 86.3 \\
\bottomrule
\end{tabular}%
}
\end{table*}

\clearpage
\clearpage

\end{document}